\documentclass[10pt,twocolumn,letterpaper]{article}

\usepackage{wacv}      

\usepackage{graphicx}
\usepackage{amsmath}
\usepackage{amssymb}
\usepackage{booktabs}
\usepackage[pagebackref,breaklinks,colorlinks]{hyperref}
\usepackage[table]{xcolor}
\usepackage{float}

\usepackage{capt-of}    

\usepackage[capitalize]{cleveref}
\crefname{section}{Sec.}{Secs.}
\Crefname{section}{Section}{Sections}
\Crefname{table}{Table}{Tables}
\crefname{table}{Tab.}{Tabs.}

\def\wacvPaperID{0000} 
\def\confName{****}
\def\confYear{2025}

\begin{document}

\title{SceneTextStylizer: A Training-Free Scene Text Style Transfer Framework \\ with Diffusion Model}

\author{Honghui Yuan\\
The University of Electro-Communications\\
Tokyo\\
{\tt\small yuan-h@mm.inf.uec.ac.jp}
\and
Keiji Yanai\\
The University of Electro-Communications\\
Tokyo\\
{\tt\small yanai@mm.inf.uec.ac.jp}
}

\twocolumn[{
    \maketitle          

    \begin{center}
      \vskip -3mm~
      \includegraphics[width=0.95\linewidth]{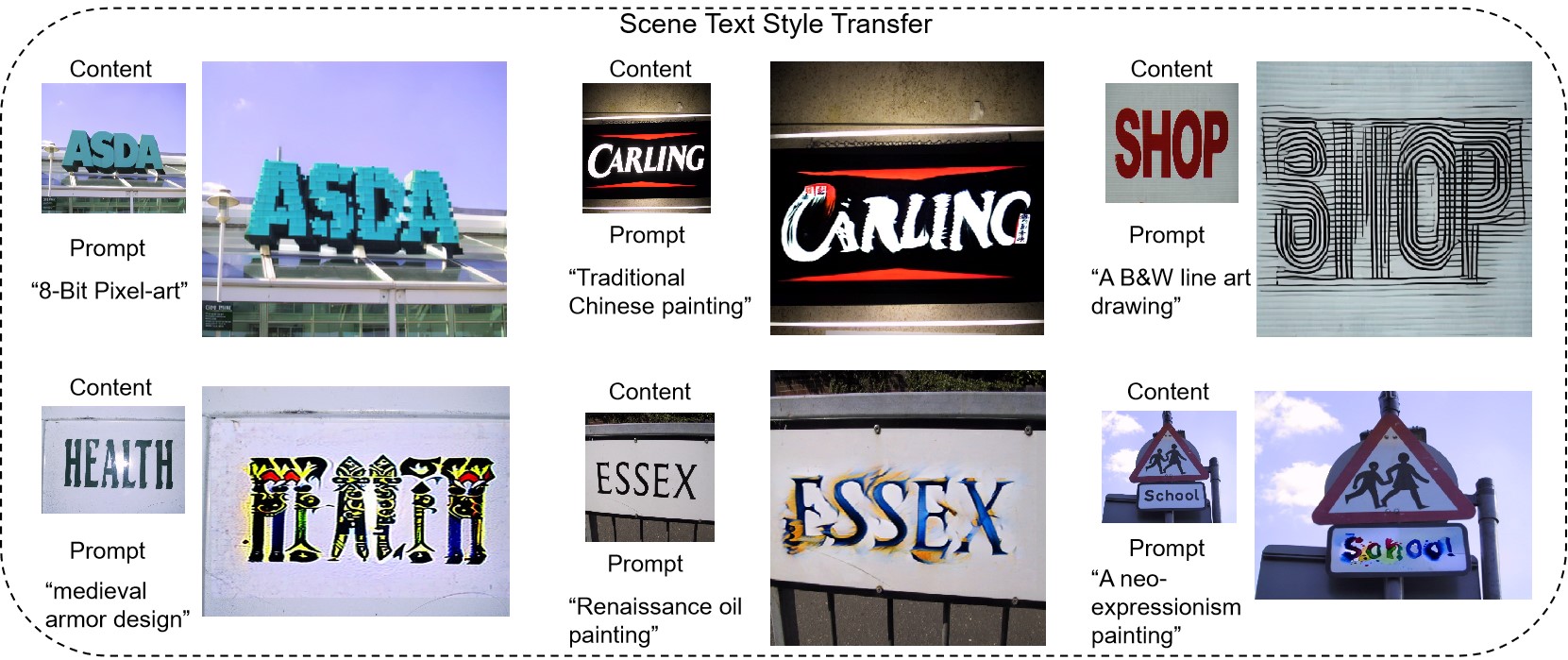}
      \captionof{figure}{Provide a scene text image and style prompt, our method can convert the text part of the image to the corresponding style of the prompt. And ensuring the background and text content remain unchanged.}
      \label{fig:top}
      \vspace{2mm}   
    \end{center}
}]

\begin{abstract}
With the rapid development of diffusion models, style transfer has made remarkable progress. However, flexible and localized style editing for scene text remains an unsolved challenge. Although existing scene text editing methods have achieved text region editing, they are typically limited to content replacement and simple styles, which lack the ability of free-style transfer.
In this paper, we introduce SceneTextStylizer, a novel training-free diffusion-based framework for flexible and high-fidelity style transfer of text in scene images. Unlike prior approaches that either perform global style transfer or focus solely on textual content modification, our method enables prompt-guided style transformation specifically for text regions, while preserving both text readability and stylistic consistency.
To achieve this, we design a feature injection module that leverages diffusion model inversion and self-attention to transfer style features effectively. Additionally, a region control mechanism is introduced by applying a distance-based changing mask at each denoising step, enabling precise spatial control. To further enhance visual quality, we incorporate a style enhancement module based on the Fourier transform to reinforce stylistic richness.
Extensive experiments demonstrate that our method achieves superior performance in scene text style transformation, outperforming existing state-of-the-art methods in both visual fidelity and text preservation.
\end{abstract}
\section{Introduction}
Style transfer has made significant progress in natural image domains. Traditional GAN-based methods typically require many style image exemplars to learn diverse stylistic features.
Recently, many diffusion-based methods~\cite{wang2023stylediffusion,huang2024diffstyler,zhang2023inversion} allow for style transfer using textual descriptions instead of explicit style images, effectively addressing the challenge of collecting large-scale style datasets. Text-guided approaches reduce data collection overhead and enable more flexible and intuitive control, especially when the desired style image is hard to obtain.
Moreover, emerging Flux-based~\cite{labs2025flux} diffusion models introduce the capability to control style transformation in specific regions using masks or textual prompts. However, accurately stylizing specific objects—especially text—remains unsolved due to the unique challenge of preserving semantic readability under stylistic changes.

For editing text regions in images, the task of Scene Text Editing (STE) has been developed to modify the textual content embedded within natural scenes. 
Traditional STE approaches~\cite{wu2019editing, yang2020swaptext} are typically limited to content replacement and not able to altering stylistic aspects of the text. Recent efforts have explored simple style modifications, such as changing the font and color.
Furthermore, some methods have trained the diffusion model with a large dataset to perform style changes by referencing stylistic cues from existing text within the same image. Although these methods support limited style transformation, they remain restricted to relatively simple or homogeneous styles.

The existing methods have the following limitations with respect to stylized transformations for text regions in the image.
\begin{enumerate}
    \item \textbf{Style matching}. 
    Existing STE methods do not support free-form style transfer. In style transfer tasks, style features are typically distributed across the entire image, and difficulty with target-specific style transfer. Text regions in scene images usually occupy only a small portion of the image and possess complex structural constraints, which makes it more difficult to convey stylistic features clearly within the text.
    \item \textbf{Text readability}. 
    Style transformation often modifies the shape, structure, or appearance of objects in the image. When applied to text, excessive stylization can severely compromise readability, especially in dense or long word sequences. Although some existing methods attempt to address this issue using character-level masks or content-aware models, they are typically effective only for one character and struggle with continuous text, particularly in realistic scene text scenarios involving longer strings or complex layouts.
    \item \textbf{Natural results}. 
    A critical objective in scene text style transfer is to ensure that the stylized text integrates naturally into the visual context, both semantically and visually. This involves not only preserving stylistic fidelity and readability but also maintaining seamless blending at the boundaries between the stylized text and the background. While some mask-based or prompt-based image editing techniques offer effective control over large, homogeneous regions, they often fail to handle the irregular shapes and fine-grained structures of text.
\end{enumerate}

In this work, we introduce SceneTextStylizer — a training-free, prompt-guided framework to stylize the textual regions of scene images. Specifically, we design a framework that leverages DDIM inversion and the self-attention features of the diffusion model to decouple and guide the generation of content and style.
Next, we introduce a changing distance mask, which is applied at each denoising step to refine the stylization process locally within the text region, enabling spatially controlled optimization.
Finally, we propose a Fourier-based style enhancement module, which extracts high-frequency components from the U-Net backbone of the diffusion model to enrich stylistic detail and improve visual fidelity.
As illustrated in Fig.~\ref{fig:top}, our framework enables text-specific style transformation within scene images, while preserving readability and visual consistency.

Our key contributions are summarized as follows.
\begin{itemize}
    \item We propose a novel training-free diffusion-based framework for scene text style editing, capable of performing prompt-guided, real-time stylization of text regions in images.
    \item We design a novel Feature Injection module specifically for text portion style transfer, and introduce a progressive distance-based control mask for localized editing and ensure seamless blending between stylized text and background.
    \item Extensive experiments demonstrate that our method outperforms existing approaches in both visual quality and stylization flexibility, effectively solving the long-standing challenge of free-form style conversion in scene text editing.
\end{itemize}

\section{Related Work}
\subsection{Arbitrary Style Transfer}
Early neural style transfer methods focused on applying style from a reference image to a content image. Neural image style transfer~\cite{gatys2016image} was the first neural style transfer method that utilizes pre-trained neural networks to achieve style transfer based on style images. AdaIN~\cite{huang2017arbitrary} enabled arbitrary style transfer by aligning the mean and variance of content images and style images.
More recently, models such as StyleGAN~\cite{karras2019style} and StyTr2~\cite{deng2022stytr2} have explored style generation using GAN~\cite{goodfellow2014generative} and Transformer~\cite{vaswani2017attention} models. These methods require style images and focus on holistic stylization.
To overcome the reliance on style exemplars, recent approaches such as CLIPstyler~\cite{kwon2022clipstyler} have leveraged vision-language models, CLIP~\cite{radford2021learning}, to enable prompt-guided style transfer. 

With the emergence of diffusion models, a new text-guided image synthesis has emerged. Many style transformation methods based on the diffusion model have achieved high-quality results.
StyleDiffusion~\cite{wang2023stylediffusion} proposed a new content-style decoupling framework and introduced a CLIP-based style decoupling loss, which realizes interpretable and controllable style transformations by explicitly extracting content information and implicitly learning supplementary style information.
InST~\cite{zhang2023inversion} proposed an inversion-based style transformation method, in which style pictures are regarded as learnable text descriptions, and style transformation is realized through the attention layer of the diffusion model.
Yang et al.~\cite{yang2023zero} achieved style transformation without the need for fine-tuning and auxiliary networks by comparing the loss of the samples generated by the pre-trained diffusion network with the patches of the original images.
Chung et al.~\cite{chung2024style} also achieved style transformation without training by replacing the keys and values of the self-attention layer of the content image with the corresponding parts of the style image during the generation process.
Diffstyler~\cite{huang2024diffstyler} designed a dual diffusion model structure that utilized text embedding to control the generation of content and style.

While these approaches achieve impressive results in whole-image stylization, they are not designed for region-specific editing, such as selectively transforming only text regions. While also inversion-based, our method targets scene text stylization with precise regional control and zero training.

\subsection{Scene Text Editing}
Scene Text Editing (STE) aims to modify the textual regions of an image while preserving the rest of the scene. Traditional methods~\cite{roy2020stefann, wu2019editing, yang2020swaptext, luo2022siman} usually divide the task into background generation, text style generation, and reintegration modules that have a complicated network structure.
Subsequent works utilize GAN to improve editing fidelity like TextStyleBrush~\cite{krishnan2023textstylebrush} and Mostel~\cite{qu2023exploring}.
Recently, several diffusion-based methods such as DiffSTE~\cite{ji2023improving}, DiffUTE~\cite{chen2024diffute}, GlyphDraw~\cite{ma2023glyphdraw}, GlyphControl~\cite{yang2024glyphcontrol}, TextDiffuser~\cite{chen2024textdiffuser} significantly advanced in scene text generation and editing. However, many of these models still exhibit style inconsistencies.
To address this, TextCtrl~\cite{zeng2024textctrl} incorporates stylistic-structural guidance into the model design as well as the integration of a Glyph-adaptive Mutual Self-attention mechanism, which improves the stylistic consistency of the text.
DARLING~\cite{zhang2024choose} improved multitasking performance for text recognition, removal, and editing by decoupling content and style features and the Multi-task Decoder.
GlyphMastero~\cite{wang2025glyphmastero} targets editing tasks with complex characters, such as Chinese, by combining local character-level features and global text-line structures.
RS-STE~\cite{fang2025recognition} integration of text recognition and editing tasks, eliminating the complexity of modeling a design with a clear separation of background style and text content, enhancing the generation ability in real-world scenarios.

In contrast to the above methods, which either focus on content modification or make limited style changes based on in-image features, our approach targets prompt-guided, free-form style transformation of scene text without altering its content. This enables flexible and diverse stylization beyond the constraints of existing font or color attributes.

\section{Proposed Method}
Our method is distinguished by three key innovations: 
(1) a training-free self-attention-based feature injection strategy for style transfer, (2) a progressive distance-based mask to ensure spatial control over text regions, and (3) a frequency-domain enhancement to preserve high-frequency stylistic textures. These modules operate in a plug-and-play manner within a pre-trained diffusion model, requiring no additional training or fine-tuning.
\subsection{Overall Framework}
\begin{figure*}[tbp]
  \centering
   \includegraphics[width=0.95\linewidth]{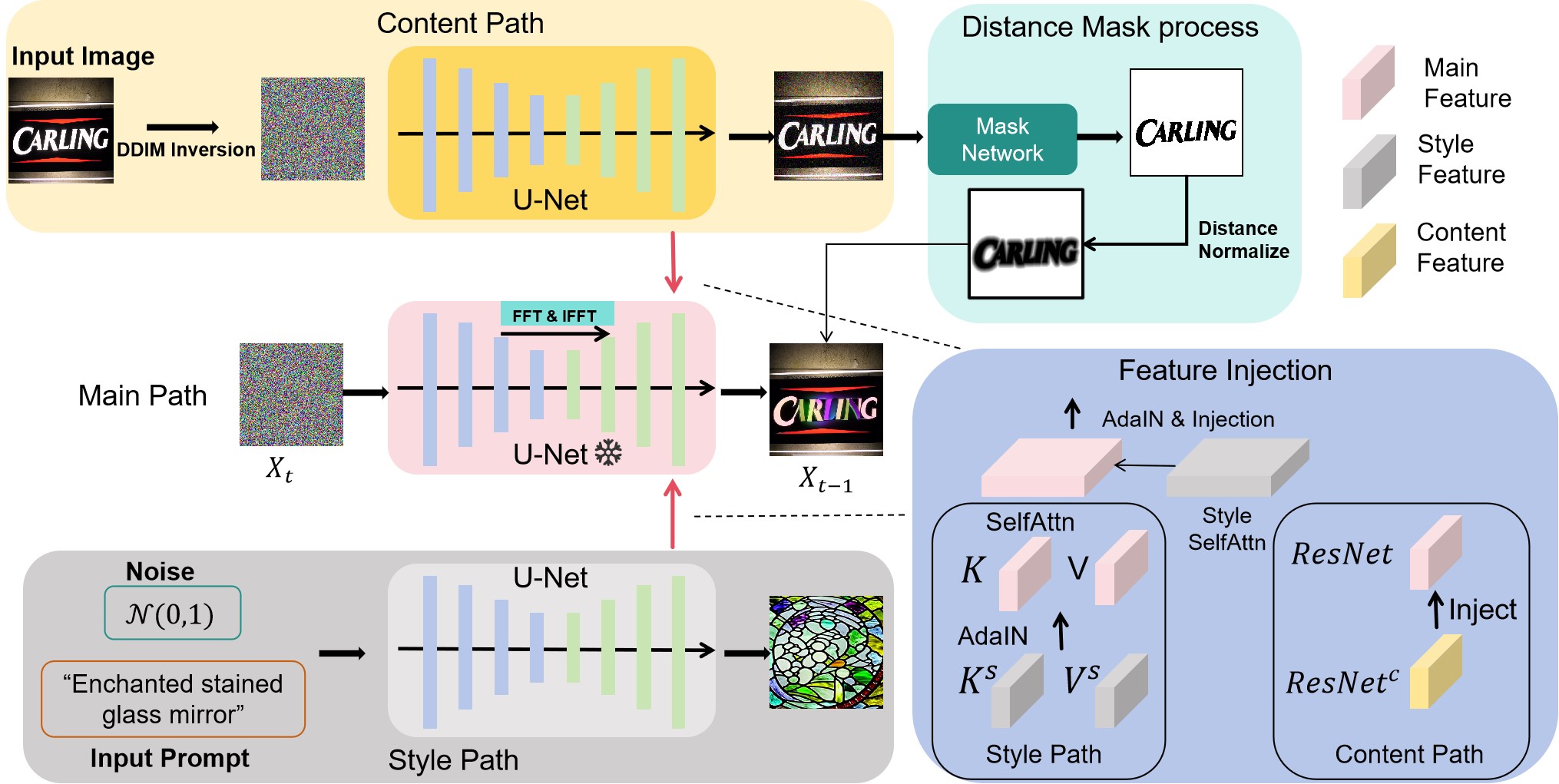}
   \caption{The framework of our method, consisting of the three denoising paths, Distance mask process, Feature Injection module targeting the text portion, and frequency module of U-Net in the main path.}
   \label{fig:net}
\end{figure*}

Given a scene text image and a style description prompt, our goal is to apply the semantic style from the prompt to the textual region of the image, while preserving the original textual content and background.
The overall structure of our framework is illustrated in Fig.~\ref{fig:net}. The framework consists of one DDIM inversion process and three forward denoising processes. We first perform DDIM inversion on the input content image to obtain its corresponding initial noise. Two denoising processes are conducted—one for the style prompt and one for the content image—to extract style and content features, respectively, via a proposed feature injection module. In the main denoising stream, these features are injected into the generation process, along with a Fourier-based style enhancement module, to synthesize the final stylized result.
To ensure region-specific control, we further introduce a distance mask that is progressively applied during denoising to constrain the stylization to the text area.

\subsection{Main Process}
The Stable Diffusion model has achieved remarkable performance in image generation tasks, primarily due to the use of attention mechanisms that effectively fuse style and content information. DDIM inversion allows mapping an image back into its corresponding noise at a specific timestep $t$, enabling controlled editing from that latent representation.

In our method, we first perform DDIM inversion on the input content image to obtain the corresponding initial noise. The content path then reconstructs the original image from this noise. In parallel, the style path uses random noise and the style prompt to generate the guided style image. The main path takes the inverted noise from the content image as input and incorporates the feature injection module, the changing distance mask, and Fourier-based enhancement during the denoising process to generate the final stylized scene text image.

\subsection{Feature Injection}
Recent training-free style transfer methods have explored using the self-attention layers of diffusion models to gain finer control over stylization. In particular, as observed in~\cite{huang2025attenst}, the Key (K) and Value (V) components in the self-attention layers play a crucial role in determining the stylistic and visual properties of the output.
As shown in Fig.~\ref{fig:net} (right), we propose a Feature Injection module. Specifically, during denoising process, we apply Adaptive Instance Normalization (AdaIN) to $K_t$ and $V_t$ tensors of the main latent representation with those extracted from the stylized image, aligning their mean and variance to ensure consistent fusion. This enables the transfer of style features into the output image. The formula is shown in the Equation~\ref{eq:kv}. 

Moreover, since style features are often weakly expressed in narrow text regions and the coherence of a long string of text. The fusion of information from other pixel positions using self-attention allows for better stylization of the text as a whole.
Thus, we inject the self-attention layer of the style path as the entire features into the main denoising step after the AdaIN process.
Furthermore, we use a step-dependent control parameter $\lambda_t$ to modulate the injection strength as shown in Equation~\ref{eq:sf}. Specifically, we apply a sigmoid schedule over the denoising steps, where low steps receive weak injection (preserving global structure), and high steps receive stronger injection (enhancing local style details).

\begin{equation}
  K_t^{\text{mix}} = \text{AdaIN}(K_s, K_t),  \ \ V_t^{\text{mix}} = \text{AdaIN}(V_s, V_t)
  \label{eq:kv}
\end{equation}

\begin{equation}
\begin{split}
  \text{Self-Attn} & (Q_t, K_t, V_t)^{\text{mix}} = \text{Self-Attn}(Q_t, K_t, V_t) \\
  &\quad + \lambda_t \cdot \text{AdaIN}(\text{Self-Attn}(Q_t, K_s, V_s))
\end{split}
\label{eq:sf}
\end{equation}

In order to ensure that the background content remains unchanged during the denoising process, inspired by Artist~\cite{jiang2024diffartist}, we also replace the hidden features of the ResBlock in the U-Net backbone with those from the content image. These features encode semantic content and help maintain the original scene layout. Through the above operation, the Feature Injection module can integrate the style features and content features in the denoising process to realize the training-free style transformation for text portions.

\subsection{Text Area Control}
The STE task typically requires large datasets of paired images and corresponding masks for training to enable text portion control. However, such data is limited or hard to obtain for style transfer tasks. Recently, Differential~\cite{levin2025differential}, which uses masks to control spatial editing in diffusion, and the strength mask enables pixel-level editing of an image. However, directly applying the masks is insufficient for text regions due to the narrow, irregular, and disconnected structures. Moreover, it often fails in transferring styles.

\begin{figure*}[tbp]
  \centering
   \includegraphics[width=1.0\linewidth]{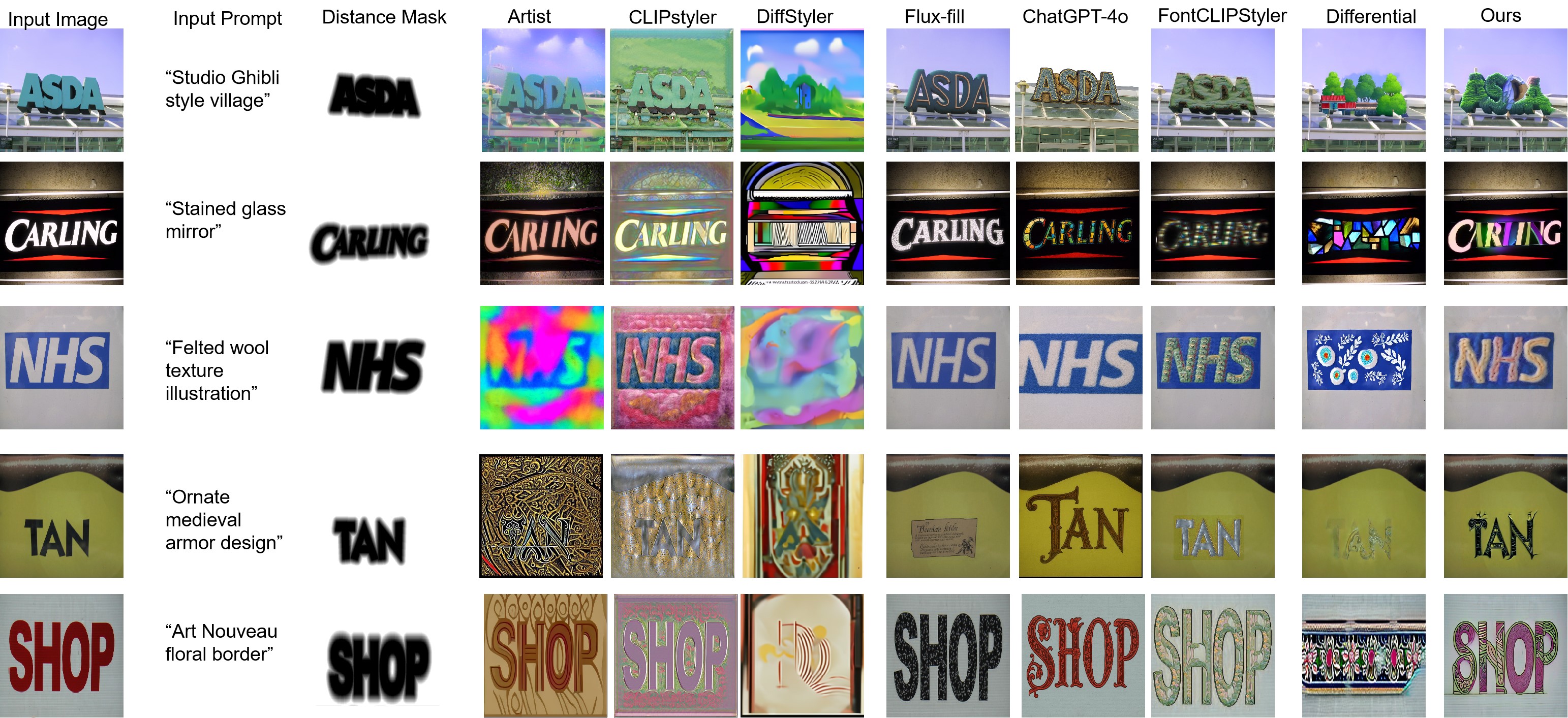}
   \caption{Qualitative comparison with state-of-the-art methods.}
   \label{fig:comp}
\end{figure*}

For text portion control without training and natural blending with the background, we propose a distance-based progressive mask injection mechanism during the denoising process. 
First, we use OCR to detect the text region of the image, and then we use the mask network provided by FontCLIPStyler~\cite{yuan2024font} to get the initial mask images. Then, a distance map that softly transitions from the center of the text (value = 1) to the background (value = 0) is calculated, capturing a gradient at text boundaries. Some results of our distance mask are shown in the third column of Fig.~\ref{fig:comp}.

During denoising, we divide the latent space into three regions: (1) the text region (random noise), (2) the integration zone (blended noise), and (3) the background (inverted content noise). At each step, we inject more style noise into the integration zone based on the distance map and timestep.
The denoised result of each step is combined with the distance mask, which is down-sampled to the latent space. This blended representation is then passed into the next denoising step. Through this process, we enable smooth spatial control of the stylization and ensure the text and background are naturally fused in the final output. 

We create the corresponding distance mask (\textit{mask\_t}) for each denoising step in all steps $T$, and we set a threshold $\left( \frac{t-T}{t} \right)$ to control the distance range for each step $t$. Equation~\ref{eq:mask} is the formula for the mask. The whole denoising process by adding the mask is shown in Equation~\ref{eq:deno}. The random noise is gradually added to the input of each denoising step according to the integration zone from the distance mask. The denoised result $z_t^{\text{mix}}$ of each step is obtained from the latent of the current timestep $z_{t+1}$ and predicted denoised latent $z_t$ combined with the mask image. Through the above operations, we can apply the style features to the text field and blend the style features and background features in the combining area to generate a natural image, and ensure the background remains unchanged.

\begin{equation}
  \textit{mask\_t} = mask \odot \left( \frac{t - T}{t} \right)
  \label{eq:mask}
\end{equation}

\begin{equation}
  z_t^{\text{mix}} = z_{t+1} \odot \text{mask\_t} + z_t \odot (1 - \text{mask\_t})
  \label{eq:deno}
\end{equation}

\begin{table*}[htbp]
  \centering
  {\small{
  \begin{tabular}{l|ccc|ccccc}
    \toprule
    Method & Artist & CLIPstyler & DiffStyler & Flux-fill & ChatGPT-4o & FontCLIPStyler & Differential & Ours\\
    \midrule
    Regional Edit  &\texttimes & \texttimes& \texttimes&\checkmark &\checkmark &\checkmark & \checkmark & \checkmark\\
    LPIPS$\downarrow$ &0.6610  & \bf 0.5986 &0.6552 &0.6794 &\underline{0.6521} &0.6776 & 0.6900 & {\cellcolor{gray!20}0.6530}\\
    DISTS$\downarrow$ & \bf 0.4377  & 0.4502 & \underline{0.4475} &0.4981 &0.4925 &0.4912 & 0.5211 &{\cellcolor{gray!20}0.4801}\\
    CLIP-Score$\uparrow$ & 0.5920  & \bf0.8088 & \underline{0.6633} & 0.4664 & 0.5181 &0.5050 & 0.5049 & {\cellcolor{gray!20}0.6070}\\
    ChatGPT-Score$\uparrow$ & 3.65  & {\cellcolor{gray!20}4.20} & 1.96 & 3.46 & \underline{4.35} &4.08 & 2.94 & \bf4.56\\
    \bottomrule
  \end{tabular}
  }}
  \caption{Quantitative evaluation between our method and previous methods, and all methods are based on the same conditions.}
  \label{tab:comp}
\end{table*}

\subsection{Improvement of Quality}
Stylizing text sequences is considerably more challenging than stylizing larger homogeneous regions due to the fine-grained and stroke-based structure of text. Often, style features are underrepresented in the output, leading to artifacts or weak stylization. 
To address this, inspired by FreeU~\cite{si2024freeu}, we enhance the U-Net structure of the diffusion model by modifying the skip-connections to better preserve and amplify style signals.

Specifically, we introduce the parameter $s$ to control the high-frequency signals in the skip connection features $f_{\text{skip}}$. In U-Net, low-frequency components govern global structure and layout, while high-frequency components control fine textures and stylistic details. We apply the Fourier Transform ($\mathcal{FT}$) and the Inverse Fourier Transform ($\mathcal{IFT}$) to manipulate these signals: setting the style amplification parameter $s$ large increases the high-frequency response for richer stylization. Because our framework already uses content inversion and mask-guided editing to preserve structure, we can safely boost high-frequency features to enhance style expression, particularly within compact text regions. The formula is shown in Equation~\ref{eq:iqu}.

\begin{equation}
  \tilde{f}_{\text{skip}} = \mathcal{IFT}\left( s \cdot \mathcal{FT}(f_{\text{skip}})) \right)
  \label{eq:iqu}
\end{equation}

\section{Experiments}
\subsection{Implement Details}
We conduct our experiments using Stable Diffusion v2.1 as the base model. The sampling processes are configured with 75 steps. The ResBlock layers used for content preservation are set to the first four layers [0, 1, 2, 3], and the injection layers for style features in the self-attention are set to 8 layers [0, 1, 2, 3, 4, 5, 6, 7]. The mask distance threshold is fixed at 5, and we set the frequency module parameter $s$ to 1.4.
All experiments are conducted on a single NVIDIA RTX 4090 GPU, with an average generation time of approximately 30 seconds per image.

\subsection{Comparison with State-of-the-Art Methods}
We compare our method with the state-of-the-art approaches that utilize prompts for control, including style transfer and regional editing methods. In addition, we also compare with multi-modal generative models ChatGPT-4o due to its robust performance for image generation capability. 
As shown in Fig.~\ref{fig:comp}, our method achieves the superior overall performance, demonstrating both effective stylization of text regions and preservation of textual structure and background. In contrast, existing methods exhibit one or more of the following limitations: inability to constrain stylization to the text area, failure to preserve text content, or weak stylization effects.

Specifically, Artist and CLIPStyler apply global style transformations across the entire image, which results in uncontrolled alterations to non-text areas and insufficient focus on the text region.
Diffstyler~\cite{huang2024diffstyler} struggles to reconstruct the correct text content, and its stylization lacks precision even when the prompt explicitly specifies text-based targets. 
Flux-fill leverages region masks to guide stylization; however, the generated results often display weak or missing style features, and in some cases, text disappearance occurs.
ChatGPT-4o results are generated under complex prompts that require the detection of textual domains and focus on the text portion style transformation. Meanwhile, keep other regions unchanged. However, its outputs still suffer from inconsistent stylization and undesired modifications to background regions, which are undesirable in style-preserving tasks.
FontCLIPstyler, although designed for text-specific stylization and achieving reasonable content fidelity, this method produced subtle and insufficient textural stylization, especially in detailed regions such as strokes and boundaries.
While Differential can utilize strength masks to achieve pixel-level editing, the results do not allow for correct text generation.

These comparisons highlight the unique advantage of our method: achieving fine-grained style transfer specifically targeted at text regions, without sacrificing background integrity or text readability.

\begin{figure*}[tbp]
  \centering
   \includegraphics[width=1.0\linewidth]{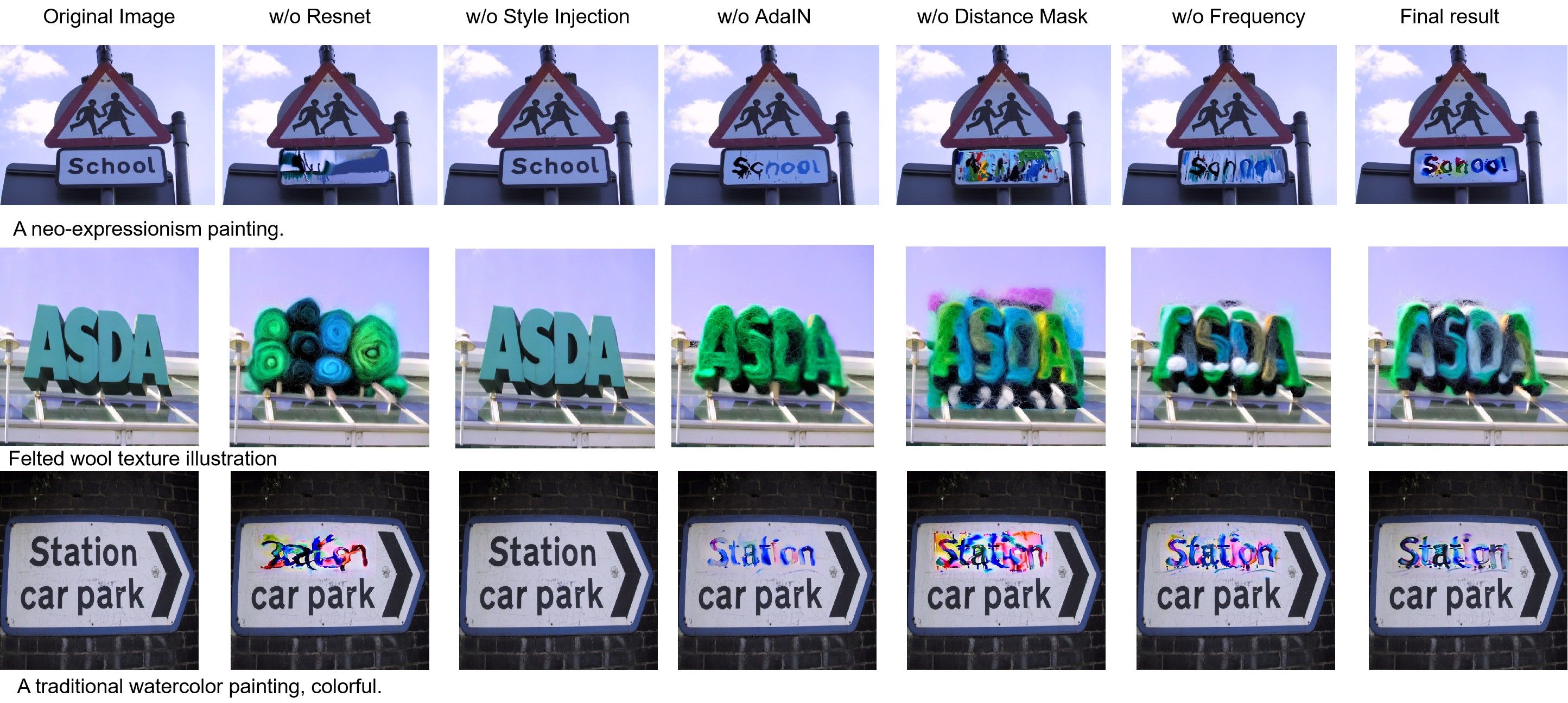}
   \caption{Qualitative evaluation results of ablation studies. The input prompts are under the images.}
   \label{fig:ab}
\end{figure*}

\begin{table*}
  \centering
  {\small{
  \begin{tabular}{l|cccccc}
    \toprule
    Method &  w/o Resnet &   w/o Style Injection &   w/o AdaIN &   w/o Distance Mask &  w/o Frequency & Final result\\
    \midrule
    LPIPS$\downarrow$ &{\cellcolor{gray!20}0.6025}  & 0.6158 &0.6143 & \bf0.5912 &0.6057 & \underline{0.6018}\\
    DISTS$\downarrow$ & \underline{0.4558}  & 0.4938 & 0.4778 & \bf0.4522 &0.4671 &{\cellcolor{gray!20}0.4564}\\
    CLIP Score$\uparrow$ & \bf 0.6356  & 0.4575 &0.5220 & \underline{0.5833} & 0.5758 & {\cellcolor{gray!20}0.5796} \\
    \bottomrule
  \end{tabular}
  }}
  \caption{Quantitative evaluation for ablation studies.}
  \label{tab:ab}
  \vspace{-3mm}
\end{table*}

\subsubsection{Quantitative evaluation}
We conduct quantitative evaluations to compare our proposed method against several state-of-the-art baselines in terms of image quality and style relevance. We randomly selected 10 content images and 15 style prompts for comparison.
To assess image fidelity and stylization quality, we adopt the Deep Image Structure and Texture Similarity (DISTS) metric, which uses the model to simulate human perception to evaluate the quality of an image in structural and textural similarity. We also employ the Learned Perceptual Image Patch Similarity (LPIPS) metric, which is a metric that calculates the similarity of texture between the generated images and the corresponding style references. For measuring alignment with the textual prompt, we use the CLIP-Score, which quantifies semantic similarity between the generated image and the input text prompt. 
To ensure consistency, we generate style reference images using Stable Diffusion v2.1 conditioned on the same prompt.
The ChatGPT-4o is used to evaluate the entire image quality. The evaluation prompt is ``Please rate the image on a scale of 0 to 5 based on the readability of the text in the image, the naturalness of the entire image, and the stylization of the text in the image.''.

The results are reported in Table~\ref{tab:comp}. The first three methods, which perform global image stylization, achieve relatively higher LPIPS scores due to strong global style transfer. However, as discussed in the qualitative evaluation results, these methods exhibit significant degradation in text readability and background preservation, leading to poor qualitative results. 
Among the other region-edit methods, our approach ranks second in LPIPS (slightly behind ChatGPT-4o), and achieves the highest DISTS and CLIP-Score, indicating superior perceptual quality and semantic consistency. Notably, our method does not require any additional training or fine-tuning, yet still outperforms most existing approaches across all metrics.
We achieved the highest ChatGPT score, proving that our images are of the highest quality.
Overall, we achieved the third-best average score among the three metrics and first in ChatGPT-score, confirming the effectiveness of our framework for scene text style editing.

\subsection{Ablation Study}
To validate the contribution of each proposed component, we conduct an ablation study by selectively removing individual modules and observing the impact on the final results. The qualitative comparisons are shown in Fig~\ref{fig:ab}.

We first evaluate the effect of the Feature Injection module. When the ResNet features from the content path are removed, the generated text loses its structural consistency and becomes unrecognizable, indicating the crucial role of content semantics in preserving textual information.
Without style injection: removing the K and V components from the style path in the self-attention layer causes the model to generate images that closely resemble the original content image, confirming their essential role for style transmission. Moreover, when we do not apply AdaIN, the stylization effect becomes significantly weaker, resulting in insufficient style expression.

We also examine the effect of the distance-based mask. When replaced with simple OCR-detected bounding boxes, the output exhibits unnatural transitions and poorly blended text-background integration. This confirms the superiority of the progressive distance mask for region-specific stylization.
Finally, we ablate the frequency module. When the module is removed, the generated images lack detailed texture. Conversely, incorporating the frequency module leads to better local texture preservation, and stylistic features are more centralized within the text. Overall, the full model—including all components—achieves the most visually pleasing and semantically consistent results, effectively incorporating rich stylistic cues into text while preserving content and natural blending.

Table~\ref{tab:ab} presents the results of the quantitative evaluation of the ablation study using LPIPS, DISTS, and CLIP-Score. The full model achieves the second and third scores of LPIPS, CLIP-Score, and DISTS, respectively. Remove ResNet, or the distance mask may yield a higher score due to broader stylization; however, these produce unnatural or distorted outputs when evaluated qualitatively. Removing AdaIN and style injection of self-attention causes greater drops in CLIP Score, suggesting that feature alignment is more crucial for semantic consistency.
Removing the Frequency module would result lower score in all metrics. 
This highlights the advantage of the full model about the trade-off between style expression and content readability, and demonstrates that each component in our framework contributes meaningfully to the overall performance.

\begin{figure}[tbp]
  \centering
   \includegraphics[width=1.0\linewidth]{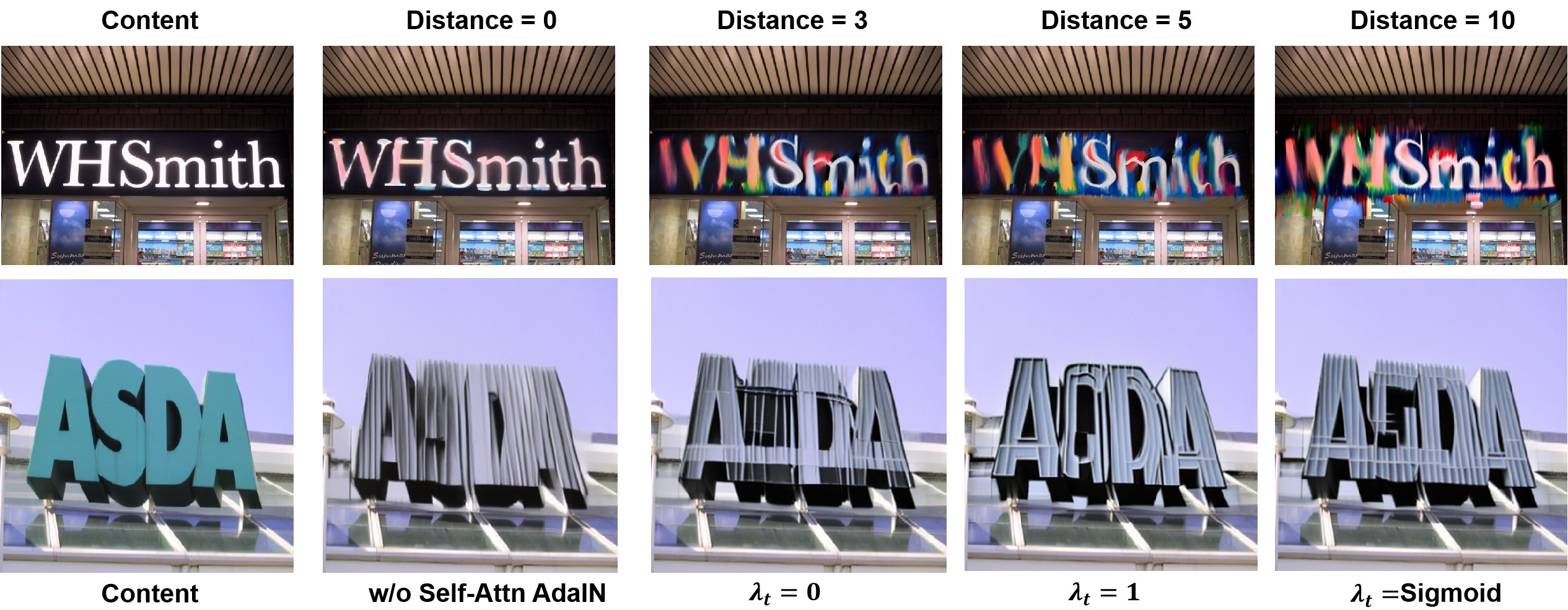}
   \caption{Results of the discussion about the distance mask
    and feature injection model. The input prompt of the first line is ``A watercolor painting'', and the second line is ``A B$\&$W line drawing''.}
   \label{fig:dist}
   \vspace{-4mm}
\end{figure}

\subsection{Discussions}
\subsubsection{Range of distance-based mask images}
Since the mask distance parameter is adjustable, we conduct a series of experiments with varying values to investigate its influence on the final results.

As shown in the first line of Fig~\ref{fig:dist}, setting the distance to 0 (using a binary mask without gradual transitions) results in poor stylized fonts, largely due to the structural complexity and narrow boundaries of text regions. When gradually increasing the distance parameter, style features are progressively diffused into the integration zone, leading to smoother and more natural transitions between text and background. However, excessively large distances (e.g., distance = 10) lead to style leakage, where stylistic features spread beyond the intended text area and disrupt visual coherence. Thus, we find that a distance value of 5 achieves a good balance, providing effective text-background blending while preserving stylistic precision.

\subsubsection{Further analysis of the Feature Injection model}
In contrast to prior methods that transform only K and V in self-attention, we perform full-layer AdaIN and introduce step-wise parameter injection specific for text portions. The second row in Fig~\ref{fig:dist} illustrates the further analysis results of our feature injection module.
In order to better display the results, the cropped text area of the generated results are displayed.
Without the AdaIN operation on the entire self-attention layer, text readability significantly degrades, highlighting its necessity for preserving structure. Furthermore, we observe that excessively large injection weight $\lambda_t$ weakens stylistic expression and with slight readability degradation. Overly small weight leads to over-stylization and also readability loss. Using the Sigmoid-based gradual fusion yields a smooth balance between content readability and style features.

\begin{figure}[htbp]
  \centering
   \includegraphics[width=0.85\linewidth]{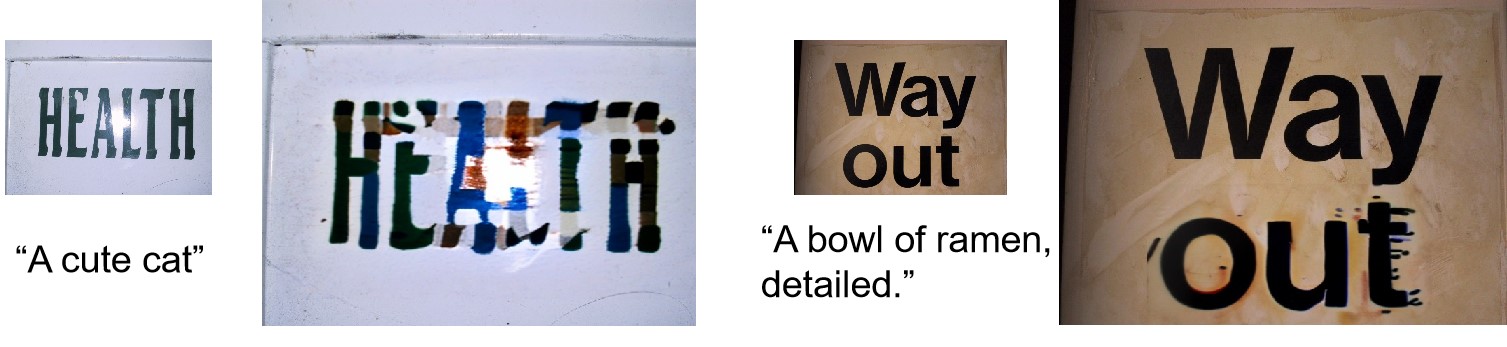}
   \caption{Some unfavorable results of our method.}
   \label{fig:lim}
   \vspace{-6mm}
\end{figure}

\subsubsection{Limitations}
Our method currently works best for texture-based or abstract styles, while object-driven prompts (e.g., "cat-style text") can lead to semantic distortion due to shape mismatch. As shown in Fig.~\ref{fig:lim}, the generated results may appear distorted or semantically inconsistent. This is largely due to the inherent difficulty in reconciling font geometry with object shape priors. And blending the semantics of the object into the constrained shape of a character is challenging for existing architectures and attention mechanisms.

While our framework performs well on texture-based and abstract styles, extending its capability to handle semantic object-based styles remains an open challenge. Future work may involve incorporating shape-adaptive representations or structure-aware diffusion strategies to better address these complex transformations.

\section{Conclusion}
In this paper, we propose SceneTextStylizer, a novel training-free diffusion-based framework for scene text style transfer guided by textual prompts. Unlike previous methods that focus on either global style conversion or limited character-level editing, our approach enables fine-grained and controllable stylization of text regions in natural images. The proposed method incorporates a self-attention-driven Feature Injection module, distance-based mask process, and frequency-domain enhancement, achieving effective text portion style transfer while maintaining text readability and background consistency. Extensive experiments validate the effectiveness of our approach, demonstrating state-of-the-art performance in scene text stylization. With zero training and prompt-based flexibility, our framework shows strong potential for personalized text design, creative media, and low-resource editing tools.

{\small
\bibliographystyle{ieee_fullname}
\bibliography{egbib}
}

\end{document}